%% file: bare_conf.tex
\documentclass[conference]{IEEEtran}
\ifCLASSINFOpdf
\else
\fi
\usepackage{amsmath}
\usepackage{amsfonts}
\usepackage{amssymb}
\usepackage{amsthm}
\usepackage{graphicx}
\usepackage{subfigure}
\usepackage{lineno}

\usepackage{pgfplots}
\pgfplotsset{compat=1.14}
\usepgfplotslibrary{fillbetween}
\usepackage{times}
\usepackage{amsthm}
\usepackage{float}
\usepackage{wrapfig}

\input{math_commands.tex}

\usepackage{fancyhdr}
\usepackage{hyperref}
\hypersetup{
     colorlinks   = true,
     citecolor    = gray
}
\usepackage{lineno}

\DeclareMathOperator{\Tr}{Tr}

\usepackage{epsfig}
\usepackage{graphicx}
\usepackage{listings}
\usepackage{amsmath}
\usepackage{amssymb}
\usepackage{tikz}
\usepackage{xcolor}
\usepackage{caption}
\usepackage{xcolor}
\usepackage[numbers]{natbib}
\usepackage{graphics}

\author{Praneeth Vepakomma, Abhishek Singh, Otkrist Gupta, Ramesh Raskar  \\
Massachusetts Institute of Technology\\
Cambridge, MA 02139, USA \\
\texttt{\{vepakom,abhi24\}@mit.edu} 
}

\begin{document}
%
\title{NoPeek: Information leakage reduction to share activations in distributed deep learning}

\maketitle


\begin{abstract}
 For distributed machine learning with sensitive data, we demonstrate how minimizing distance correlation between raw data and intermediary representations reduces leakage of sensitive raw data patterns across client communications while maintaining model accuracy.  Leakage (measured using distance correlation between input and intermediate representations) is the risk associated with the invertibility of raw data from intermediary representations. This can prevent client entities that hold sensitive data from using distributed deep learning services. We  demonstrate that our method is resilient to such reconstruction attacks and is based on reduction of distance correlation between raw data and learned representations during training and inference with image datasets. We prevent such reconstruction of raw data while maintaining information required to sustain good classification accuracies.
\end{abstract}


%

\section{Introduction}
Data sharing and distributed computation with security, privacy and safety have been identified amongst important current trends in application of data mining and machine learning to healthcare, computer vision, cyber-security, internet of things, distributed systems, data fusion and finance. \cite{aditya2016pic,orekondy2017towards,schiff2009respectful,frome2009large,yu2007privacy,orekondy2017towards,upmanyu2009efficient,senior2005enabling,avancha2012privacy,halperin2008security}.  
Hosting of siloed data by multiple client (device or organizational) entities that do not trust each other due to sensitivity and privacy issues poses to be a barrier for distributed machine learning. This paper proposes a way to mitigate the reconstruction of raw data in such distributed machine learning settings from culpable attackers. Our approach is based on minimizing a statistical dependency measure called distance correlation \cite{szekely2007measuring,sejdinovic2013equivalence,szekely2009brownian,szekely2013energy,dueck2014affinely} between raw data and any intermediary communications across the clients or server participating in distributed deep learning. We also ensure our learnt representations help maintain reasonable classification accuracies of the model, thereby making the model useful while also protecting raw sensitive data from reconstruction by an attacker that can be situated in any of the untrusted clients participating in distributed machine learning. 
\par
\begin{figure}
  \centering
  \includegraphics[scale=0.041]{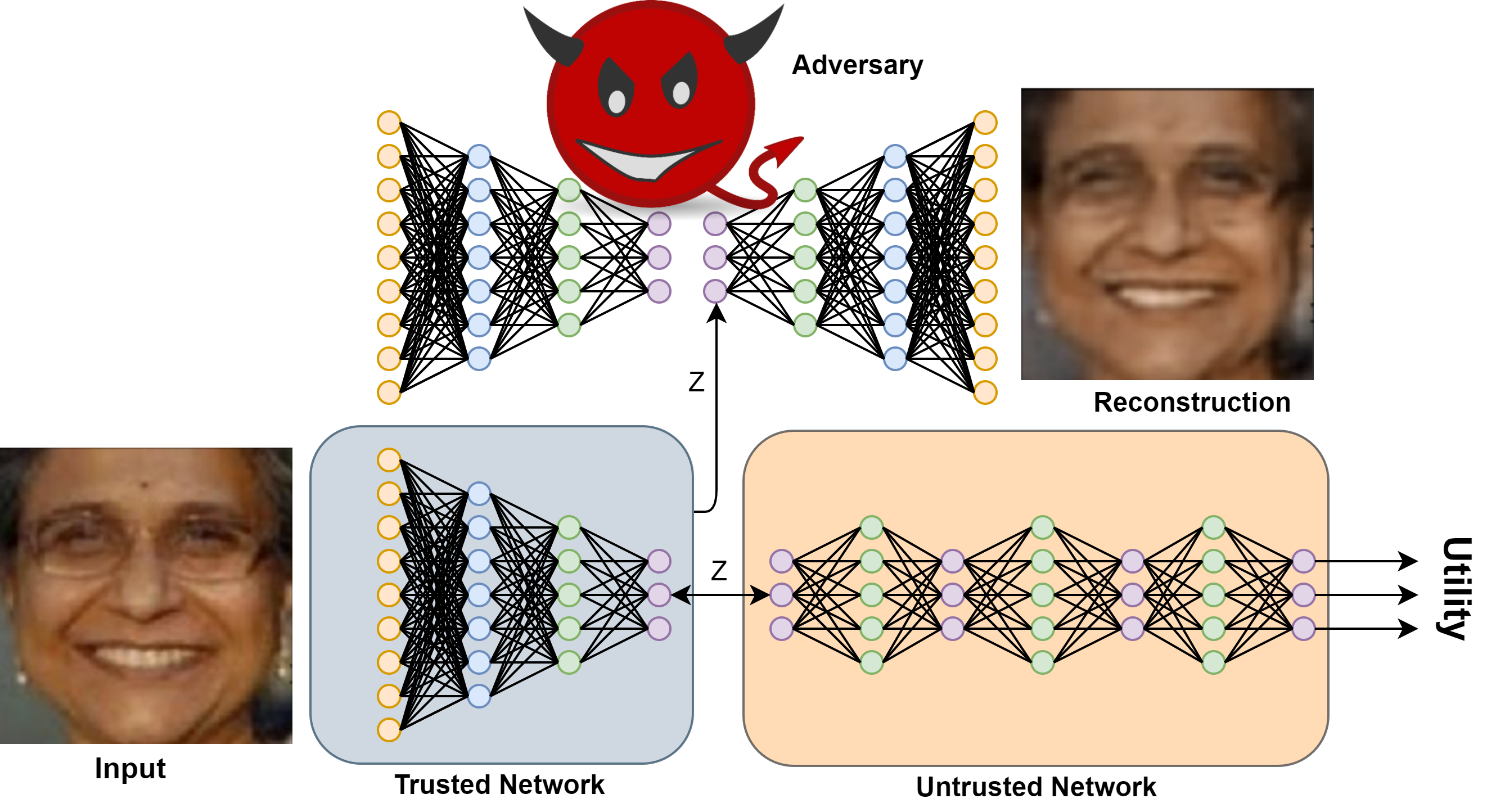}
  \includegraphics[scale=0.041]{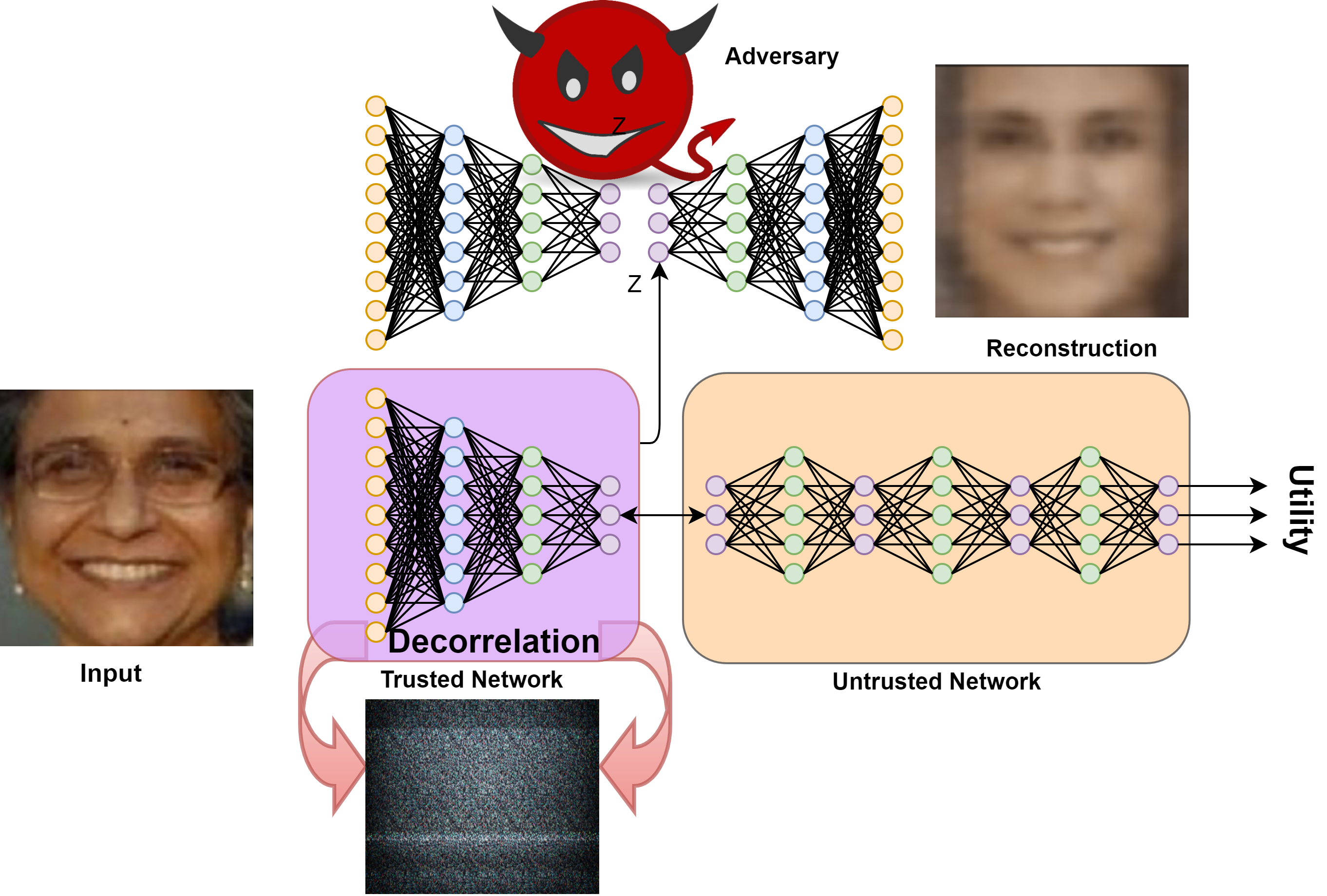}
  \caption{\textcolor{cyan}{Left:} In the regime of regular training of deep neural networks, information about sensitive raw input data is leaked through intermediate activations even after input data passes through multiple layers. As shown in this figure, upon sending intermediate activations from a trusted network on a client to an untrusted network for computing rest of the task, an adversary on server-side can reconstruct original raw data from the activations. \textcolor{cyan}{Right:} NoPeek is a method where intermediate activations are decorrelated with raw input data while training the network to obtain high classification accuracy on the untrusted network. In this figure, unlike regular deep neural network training, the adversary is not able to reconstruct the exact raw image of the person.}
  \label{fig1}
\end{figure}

 \textbf{Reconstruction attack setting:} The proposed solution aims to give greater protection to the distributed learning ecosystem from reconstruction attacks (reconstruction of raw data from transformed activations) from attackers residing in any client or server that receives communications from another client. It also protects reconstruction attacks from insider threats (attacker resides inside the client/server that transforms the sensitive data). The attack is also illustrated in Figure \ref{fig1} with regards to the regular training of deep neural networks. We now describe the popular reconstruction attack setting in greater detail along with its relevance to current real-world distributed deep learning prospects. \par \textbf{Attack assumptions:} We consider providing security in relatively worst-case settings where the attacker is given an advantage in terms of the assumptions made. This is considered to be a good practice in the community of privacy preserving machine learning as it also enables provision of security under a wider variety of plausible modifications of attack schemes with assumptions that are weaker than the assumed worst-case attacker's capacities. This level of protection is thereby expected to be offered by a working solution in addition to its value in the worst-case setting assumed. In worst-case reconstruction attack settings, the attacker has access to a leaked subset of  samples of training data along with corresponding transformed activations at a chosen layer, the outputs of which are always exposed to other clients/server by design for the distributed training of the deep learning network to be possible. The attacker could reside in any untrusted client or server that is part of the distributed training setup. The attacker also has access to rest of the activations corresponding to unleaked training data at the same layer. This is also by design, in order for the distributed training to be functionally possible. The attacker tries to learn an image to image translation model from the transformed activations to the leaked raw data. The attacker can then use that model to reconstruct raw data from activations corresponding to unleaked training data or unleaked test/validation data by inferring from the learnt reconstruction model that was trained on corresponding pairs of activations and raw samples of leaked data.
\par \textbf{Attack implications:} Typically leakage of a sub-sample of raw data has serious financial, ethical, legal, public relation (PR) and regulatory implications. Such leakages have continued to happen in recent times and often the ratio of $\frac{\text{\# of records leaked}}{\text{total \# of records owned by the institution}}$ is quite small and yet the real-life negative implications of such a leak are massive. According to `2019 Cost of a Data Breach Report' in \cite{dataBreach}, the cost per data breach is between \$$1.25$ million  to \$$8.19$ million depending on country and industry at which the breach occurs. The average size of a data breach is established to be around $25,575$ records.  The total global cost of data breaches runs into billions of dollars per year. In addition $60$\% of small and medium enterprises (SMEs) that experience a cyber breach go out of business in the following $6$ months according to an official government report in \cite{dataBreach2}. The lifecycle of a typical data breach is estimated to be $279$ days and for that of a malicious attack is estimated to be $314$ days.  The goal henceforth is to prevent the attacker from using the leaked data to construct an inverse model that can reconstruct other raw data records upon just looking at the communicated intermediate activations received at the server as is required by the important distributed deep learning settings cited above. 
\par \textbf{Relevance of attack setting:} We describe two popular settings of distributed deep learning where this attack setting is highly relevant. 
\begin{enumerate}
    \item \textbf{Split Learning:} This attack model is highly relevant to a popular resource and communication efficient variant of distributed deep learning called split learning \citep{gupta2018distributed,vepakomma2018split,singh2019detailed}. In this setting, intermediate activations from a chosen layer (called split leayer) of the deep network  are communicated from client to the server during training. The rest of the network is processed at the server during forward propagation. In turn, during backpropagation the gradients from the server's first layer (layer next to the split layer) are communicated back to the client. The rest of backpropagation occurs at the client. These rounds of communication are continued to finish all the epochs of distributed training. Split learning has also been ported into PySyft by OpenMined, a popular and widely adopted opensource framework for privacy preserving distributed machine learning.  Split learning has been adopted in Internet of Things (IOT) and edge-device machine learning settings in  \citep{koda2020communication,gao2020end,thapa2020splitfed,lim2020incentive} including multi-modal fusion based machine learning across edge devices in \citep{koda2019one, park2020extreme} with data collected at the edge on imagery and millimeter wave (mmWave) radio frequency (RF) signals to perform distributed machine learning. Split learning has also been used for Ultra Reliable Low Latency Communication (URLLC) settings with distributed learning as part of 5G communication research.  Suitability of split learning for healthcare has been described in \cite{allen2019democratizing}. The work in \cite{mireshghallah2019shredder} considers protection of intermediate activations under this attack setting solely for private inference via learning of specific noise distributions to perturb the activations prior to communication. Our work instead focuses on training. 
    
    \item \textbf{Adversarial reconstruction:} This threat model has also been considered with regards to adversarial reconstruction attack settings such as those considered in \cite{yang2019adversarial} based on activations obtained at the end of neural network. Server-side insider threats that aim to reconstruct raw data of the client are another realistic example of this attack setting.  \cite{li2019deepobfuscator} attempts to learn activations of a given network at chosen layers while attempting to protect an adversarial reconstructor that attempts to reconstruct entire raw data or partial attributes of raw data from these activations. 
\end{enumerate}

The attack settings outside the purview of reconstruction attacks that we do not consider in this paper include those of model extraction, model inversion, malicious training, adversarial examples (evasion attacks) and membership inference.

\subsection{Contributions}
We show that reducing the distance correlation between learned representations and raw data prevents information leakage with regards to sensitive data in machine learning settings. This is illustrated in Figure \ref{fig1}. The decorrelated data can then be used for various machine learning tasks as long as it holds enough information to perform the intended task while not having enough information to reconstruct the raw data itself. Our developed methods apply to the following settings:
\begin{enumerate}
    \item Training schemes to prevent reconstruction of entire raw data or specifically chosen attributes in deep learning during inference.
    \item Device-level sanitization as burn-in period to reduce leakage of information during the initial epochs of training while not requiring the client that holds the raw data to communicate with the server. Following this burn-in period, the client and server entities train with communication between them.
    \end{enumerate} 
We evaluate the method and share detailed results via a reconstruction testbed we describe in the experiments section. 
\subsection{\textbf{Code/Reproducibility}} The code for our method is provided in this anonymous code repository: \url{https://anonymous.4open.science/r/820473f8-f3ee-4212-9b9c-409a78722af6/}. We will also release seeds, trained models, training logs, and intermediate data files for improving reproducability.

\section{Related Work}
We primarily focus on the modality of image/computer vision datasets to analyze and test our proposed method. To maintain specificity we broadly categorized related works on security and privacy for this modality as follows.
\paragraph{Deep learning, adversarial learning and information theoretic loss based privacy} 
These can be categorized into hiding specific sensitive attributes using adversarial training that reaches an equilibrium based on information theoretic loss functions optimized under minimax settings through learning weights of deep learning models \citep{li2019deepobfuscator, mirjalili2019flowsan, roy2019mitigating, zhang2018mitigating, edwards2015censoring,wu2018towards,xu2019ganobfuscator}. A kernelized version of such an adversarial learning approach with theoretical guarantees is provided in \cite{sadeghi2019global}. A similar non-adversarial approach that still uses a dependency measure of maximum mean discrepancy (MMD) for learning a variational autoencoder is in 
\cite{louizos2015variational,zemel2013learning}. The method we propose in our paper is not necessarily tied to a generative adversarial network (GAN) styled architecture where two separate models have to be trained in tandem. Our proposed model is based on a easily implementable differentiable loss function between the intermediate activations and the raw data. This will be described in detail later on in section \ref{metheodsec}.
\paragraph{Homomorphic encryption and secure multi-party computation for computer vision:}
    Homomorphic encryption (HE) and multi-party computation (MPC) techniqes although highly secure are not computationally scalable and communication efficient for complex tasks like training large deep learning models. Thereby their application to machine learning has been with regards to smaller computations or specific applications requiring computation of functions with a much simpler complexity. The work in \cite{bringer2013privacy,boddeti2018secure,phong2018privacy} describes HE and garbled circuits (an MPC scheme) for privacy-preserving biometric identification. Similarly \cite{avidan2006blind, bringer2014gshade} uses MPC schemes like oblivious transfer, secure millionaire, secure dot product for face matching with respect to a collection of sensitive surveillance footage images. \cite{yonetani2017privacy} uses HE for secure aggreagation of classifiers in a distributed learning setting where different entities train models on their local data and share the model weights or model related information that needs to be securely aggregated at a centralized entity. Our proposed method in this paper is communication efficient and highly scalable computationally with regards to large deep learning architectures for both training as well as inference attacks unlike the HE and MPC based models.
    \paragraph{Differential privacy for computer vision:} Differential privacy schemes are based on adding noise dependent on the query under computation to primarily provide privacy against membership inference attacks. The works in \cite{hitaj2017deep,wang2015differentially,papernot2016semi} are examples of such schemes for transfer learning, subspace clustering. A modified scheme called separated-DP \cite{bhowmick2018protection} aims to provide guarantees against reconstruction in the context of federated learning \cite{konevcny2016federated}, a popular distributed learning method for the purpose of protecting model weights trained at individual client entities while securely aggregating the average of these weights at a centralized server. These methods typically take a stronger hit on accuracy of deep learning models although at the benefit of attempting to provide worst-case privacy guarantees for membership inference attacks.
    

\section{Method}\label{metheodsec}

 \begin{figure}[h]
 \includegraphics[scale = 0.65]{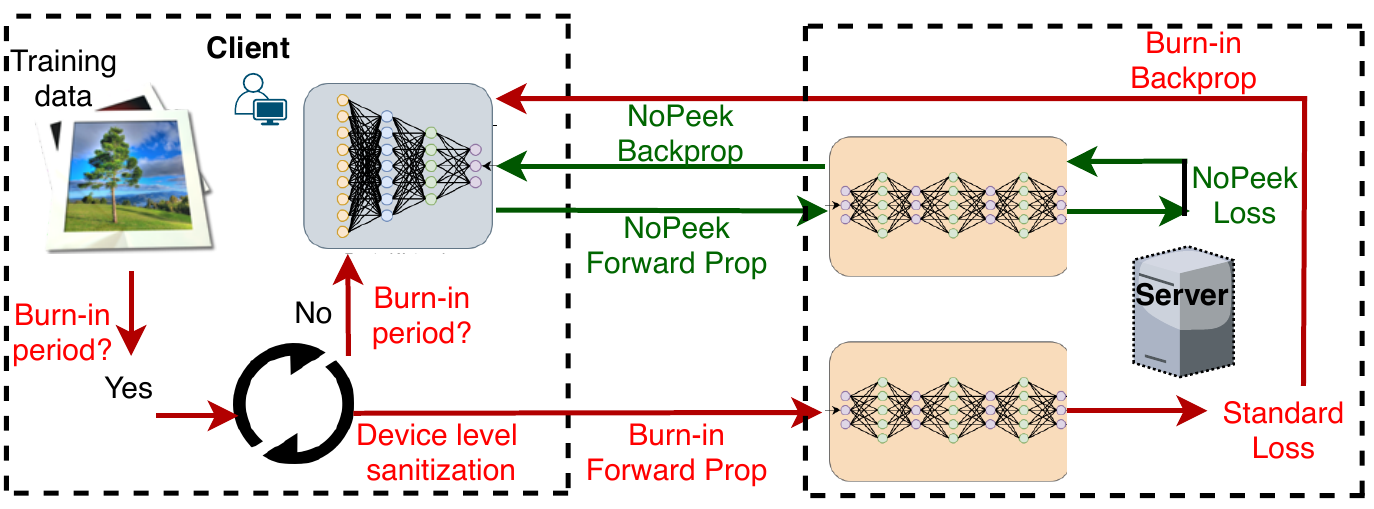}
 \caption{Workflow for NoPeek method including a) device-level santization on the client during a burn-in period (in red) followed by b) training with the NoPeek method (in green) to prevent reconstruction of data while maintaining classifcation accuracies. There is no communication between client (left) and server (right) during the burning period until the device-level sanitization is completed.}
 \label{UnivSmasherPipeline}
 \end{figure}
 

\begin{figure}
    \centering
    \includegraphics[scale=0.35]{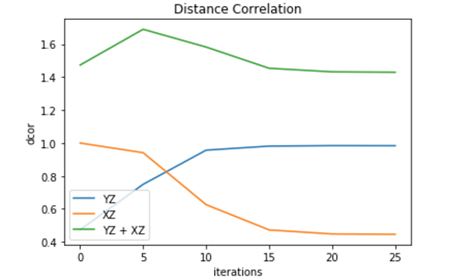}
    \caption{Universal decorrelation iterations show reduction in distance correlation with raw data (orange) while preserving distance correlation needed to complete the task (blue) on CIFAR-10 data. This scheme is useful to reduce leakage as burn-in period prior to starting distributed deep learning as this does not require any communication with the outside network. This scheme is not required during inference any more as decorrelator is trained by then. We observe in our experiments that this is crucial in preventing reconstruction during initial epochs of distributed training post the burn-in period.}
    \label{fig:universal_smasher_cifar}
\end{figure}

\textbf{Key idea:} The key idea of our proposed method is to reduce information leakage by adding an additional loss term to the commonly used classification loss term of categorical cross-entropy. The information leakage reduction loss term we use is distance correlation; a powerful measure of non-linear (and linear) statistical dependence between random variables. The distance correlation loss is minimized between raw input data and the output of any particularly chosen layer whose outputs need to be communicated from the client to another untrusted client or untrusted server. Optimization of this combination of two losses helps ensure the activations resulting from the protected layer have minimal information with regards to reconstructing the raw data while still being useful enough to achieve reasonable classification accuracies upon post-processing of these activations. The quality of preventing reconstruction of raw input data while maintaining reasonable classification accuracies is qualitatively and quantitatively substantiated in the experiments section. Therefore, layers from the raw data upto the protected layer act as decorrelation layers that preserve classification utility.\par \textbf{Loss function:} The total loss function for $n$ samples of input data $\mathbf{X}$, activations from protected layer $\mathbf{{Z}}$, true labels $\mathbf{Y}_{true}$, predicted labels ${\mathbf{Y}}$ and scalar weights $\alpha_1,\alpha_2$ is therefore given by \begin{equation}
    \alpha_1 DCOR(\mathbf{X,Z}) + \alpha_2 CCE(\mathbf{Y}_{true},\mathbf{Y})
\end{equation}

The gradient of distance correlation is provided below for optimization purposes in Appendix \ref{gradDC} although we optimize the above loss function using \textit{Autograd} as we use it in the context of distributed deep learning. A deep learning friendly code for computing distance correlation is also provided there. A link to our anonymous code repository is also provided there for reproducibility.
\begin{wrapfigure}[23]{r}{0.21\textwidth}
    \centering
    \includegraphics[scale=0.25]{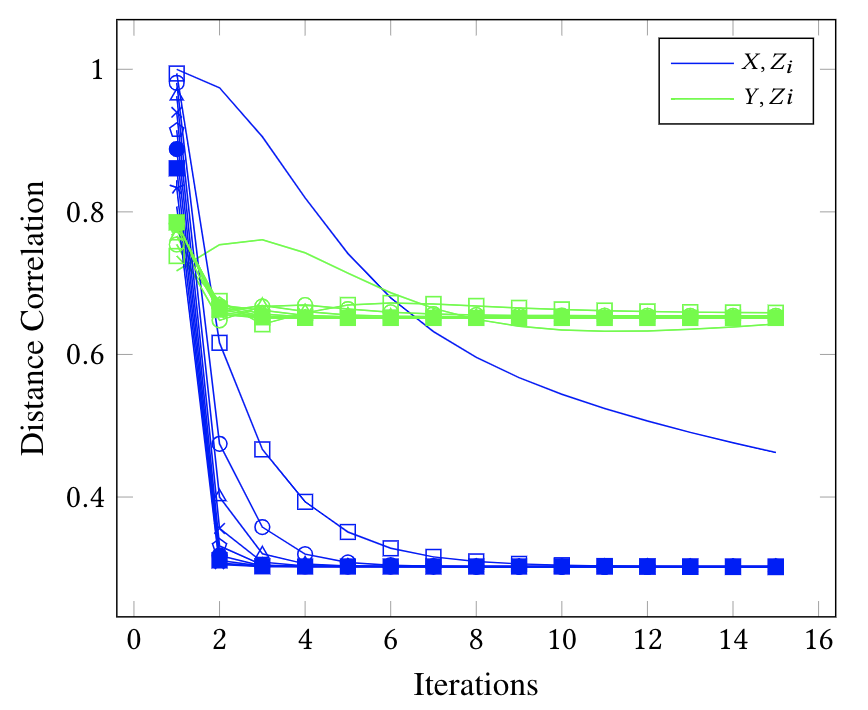}
    \caption{Universal decorrelation iterations show reduction in distance correlation with raw data (blue) while preserving distance correlation needed to complete the task (green) on Boston Housing data. This scheme is useful to reduce leakage as burn-in period prior to starting distributed deep learning as this does not require any communication with the outside network.}
    \label{fig:universal_smasher_cifar1}
\end{wrapfigure}

\subsection{Initialization with device-level decorrelation}
\begin{figure}
    \centering
    \includegraphics[scale=0.12]{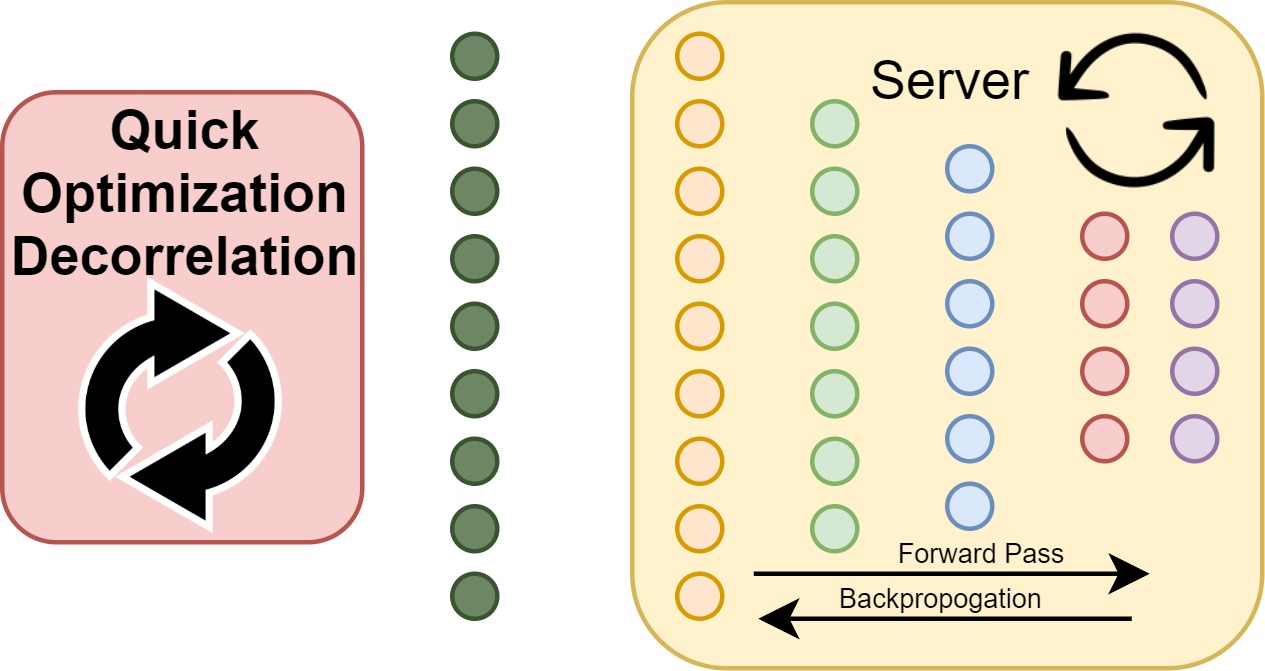}
    \caption{The device level sanitization scheme helps to control leakage during early iterations of model training  by acting as a burn-in period to initialize the activations for the deep learning based decorrelator. Following this the distributed deep learning model is trained. During inference, the distributed deep learning model is used directly as the model remains trained at that point of time.}
    \label{fig:universal_smasher}
\end{figure}

All iterations of training require communication between the client/server entities involved in distributed deep learning. In order to ensure that there is no leakage during the initial iterations of minimizing our proposed loss function, we perform a device-level decorrelation routine during an initial burn-in period of iterations before allowing any communication.
Therefore this is a highly communication-efficient approach for learning decorrelated representations of client's raw data
that can then be shared with other entities in a distributed learning setting. Following the burn-in period, the server receives the decorrelated activations and continues to sync with the client to perform distributed training of our loss function proposed in previous subsection. This process is illustrated at a high-level in Figures \ref{UnivSmasherPipeline}, \ref{fig:universal_smasher_cifar}, \ref{fig:universal_smasher_cifar1} and \ref{fig:universal_smasher}. \par Unlike traditional distributed deep learning approaches, this iterative approach does not require any gradient backpropagation or exchange of activations with outside network to perform the optimization and is particularly suitable for on-device (client) decorrelation of data prior to performing any major communications across the distributed entities.  This scheme is useful to
reduce leakage as burn-in period
prior to starting distributed deep
learning as it does not require
any communication with the outside network. Following this after the distributed deep learning based decorrelator is trained, this scheme is not required during inference any more as decorrelator is trained
to be optimal by that point of time. This approach of solely optimizing on client to initialize the distributed deep learning model with a decorrelated representation prior to beginning model training as illustrated in Figure \ref{fig:universal_smasher}. \par The idea is based on a modification of a scheme for maximizing sum of distance correlations proposed in \cite{vepakomma2018supervised}  to obtain low-dimensional representations that preserve a high distance correlation with labels $\mathbf{Y}$. Motivated by their setup which also seems similar in principle to the information bottleneck method, we instead consider a \textit{difference} of these distance correlations instead of a sum as unlike their method which was for supervised dimensionality reduction we would like to minimize distance correlation with raw data while preserving distance correlation with labels. This objective function can be expressed as below as distance correlation can be expressed in terms of specific graph Laplacians whose exact form is detailed in \cite{vepakomma2018supervised}:
\begin{align}
f(\mathbf{Z}) & = 	 \frac{
			\Tr{\mathbf{Z}^T\mathbf{L}_\mathbf{y}\mathbf{Z}}
		}{\sqrt{
			\Tr{\mathbf{Y}^T\mathbf{L}_\mathbf{Y}\mathbf{Y}}
			\Tr{\mathbf{Z}^T\mathbf{L}_\mathbf{Z}\mathbf{Z}}
		}}-\frac{
			\Tr{\mathbf{Z}^T\mathbf{L}_\mathbf{X}\mathbf{Z}}
		}{\sqrt{
			\Tr{\mathbf{X}^T\mathbf{L}_\mathbf{X}\mathbf{X}}
			\Tr{\mathbf{Z}^T\mathbf{L}_\mathbf{Z}\mathbf{Z}}
		}}
\nonumber
\end{align}

The iterative update for maximization of the above objective is based on a variant of majorization-minimization \cite{mairal2015incremental,mairal2013stochastic} and is given by $\mathbf{Z}_t = \mathbf{H}\mathbf{Z}_{t-1}$  where \[ \mathbf{H}=\left(
			\gamma ^2\mathbf{D}_X -\alpha \mathbf{S}_{\mathbf{X},\mathbf{Y}}
		\right)^{\dagger}
		(\gamma ^2\mathbf{D}_X-\mathbf{L}_{\mathbf{M}}) \]
for a fixed $\gamma^2$, some $\alpha$ and  where $k_X=\frac{1}{\sqrt{\Tr{\mathbf{X}^T\mathbf{L}_\mathbf{X}\mathbf{X}}}}$,   $k_Y=\frac{1}{\sqrt{\Tr{\mathbf{Y}^T\mathbf{L}_\mathbf{Y}\mathbf{Y}}}}$ are constants, and $\mathbf{S}_{\mathbf{X},\mathbf{y}}=k_Y\mathbf{L}_{\mathbf{Y}} -\beta k_X \mathbf{L}_{\mathbf{X}}$ for a tuning parameter $\beta$, where the details of all these parameters are also in \cite{vepakomma2018supervised} except for $\beta$ which has been added to study its effect on convergence rate as seen in Figure 2. The only other difference in the iterative updates ends up in the definition of $\mathbf{S_{XY}}$ where in the case of the supervised dimensionality reduction usecase, the negative instead becomes a positive.

\subsection{Advantages of using distance correlation}\label{motivsec}
Estimation of classical information theoretic-measures as used in \citep{mirjalili2019flowsan,roy2019mitigating, zhang2018mitigating, edwards2015censoring,wu2018towards} is a known hard problem. Recent approaches to estimate it effectively like \citep{belghazi2018mine} are based on iterative optimization. A recent data efficient version of it requires 3 nested for loops of optimization \cite{lin2019data} to estimate it. Therefore in the context of deep learning, every epoch of learning the weights is dependent on this iterative optimization. In contrast our approach uses distance correlation, a measure of non-linear (and linear) dependency that can be estimated in closed-form. Fast estimators of distance correlation require $\mathcal{O}(nlogn)$ \citep{chaudhuri2019fast,huo2016fast} computational complexity for univariate and $\mathcal{O}(nKlog n)$ complexity \cite{huang2017statistically} for multivariate settings with $\mathcal{O}(max(n, K))$ memory requirement, where
$K$ is the number of random projections. Distance correlation has been shown to be a simpler special case of other recent popular measures of dependence such as Hilbert-Schmidt Independence Criterion (HSIC), Maximum Mean Discrepancy (MMD) and Kernelized Mutual Information (KMI) that have been extensively studied and used recently in the machine learning and statistics community \citep{sejdinovic2013equivalence,tonde2016supervised,sejdinovic2012hypothesis,chang2013canonical}.  An advantage of using a simpler alternative is that in addition to it being differentiable and easily computable with a closed-form, it requires no other tuning of parameters and is self-contained unlike HSIC, MMD and KMI that depend on a choice of separate kernels for features as well as labels along with their respective tuning parameters. 

\section{Experiments}
\textbf{Reconstruction attack testbed:}
We empirically examine the privacy aspects of our method by designing a testbed which performs feature inversion. The idea of this testbed is to emulate the attacker and to evaluate the quality of reconstruction attack.
 The testbed itself is a neural network with a decoder architecture where the layers are composed of transpose convolutions. Similar architecture have been used in generative models for generating images from low-dimensional latent codes. We use other standard components like ReLU activation, batch normalization and ResNet style of performing additive skip connections. Input to this testbed is the intermediate activations, $z_{l}$ from any arbitrary layer $l$ of the target model and output is the image generated $\hat{x}$. We first train two separate ResNet-18 architectures on datasets for image classification with NoPeek and without NoPeek for baseline comparison. After the training, we use held-out validation set to generate intermediate activations. We thereby generate a paired dataset of activations and corresponding images. We use this paired dataset to train the reconstruction testbed to emulate the attacker. We use 90\% of the original validation dataset as training dataset for the reconstruction testbed and remaining 10\% is used as test-set for the qualitative evaluation of the reconstruction quality. We train the model in the reconstruction testbed on a dataset of $z_{l}, x$ pairs with the loss function as the euclidean norm between $x$ and $\hat{x}$. We want to emphasize that there can be a potentially better design for architectures of the reconstruction testbed and better loss functions but the goal of this paper is to just have a fair comparison between the NoPeek based training and the regular training of deep networks using a reasonable reconstruction architecture. The number of upsampling layers in the architecture of the testbed vary depending upon the difference in the dimensionality of $z_{l}$ and $x$.

 For all of our reported experiments for training the network, we use Adam optimizer with initial learning rate as $1\times e^{-3}$ with exponential decay.
The experiments are further detailed below.
\subsection{CIFAR10}
We use CIFAR10 for our inversion experiments with respect to vision model trained for image classification task. We first train the network on 50,000 training samples of CIFAR10 with and without NoPeek and then use 10,000 validation data samples and their corresponding activation as the dataset for the inversion attack model.
\begin{figure}
    \centering
    \includegraphics[scale=0.10]{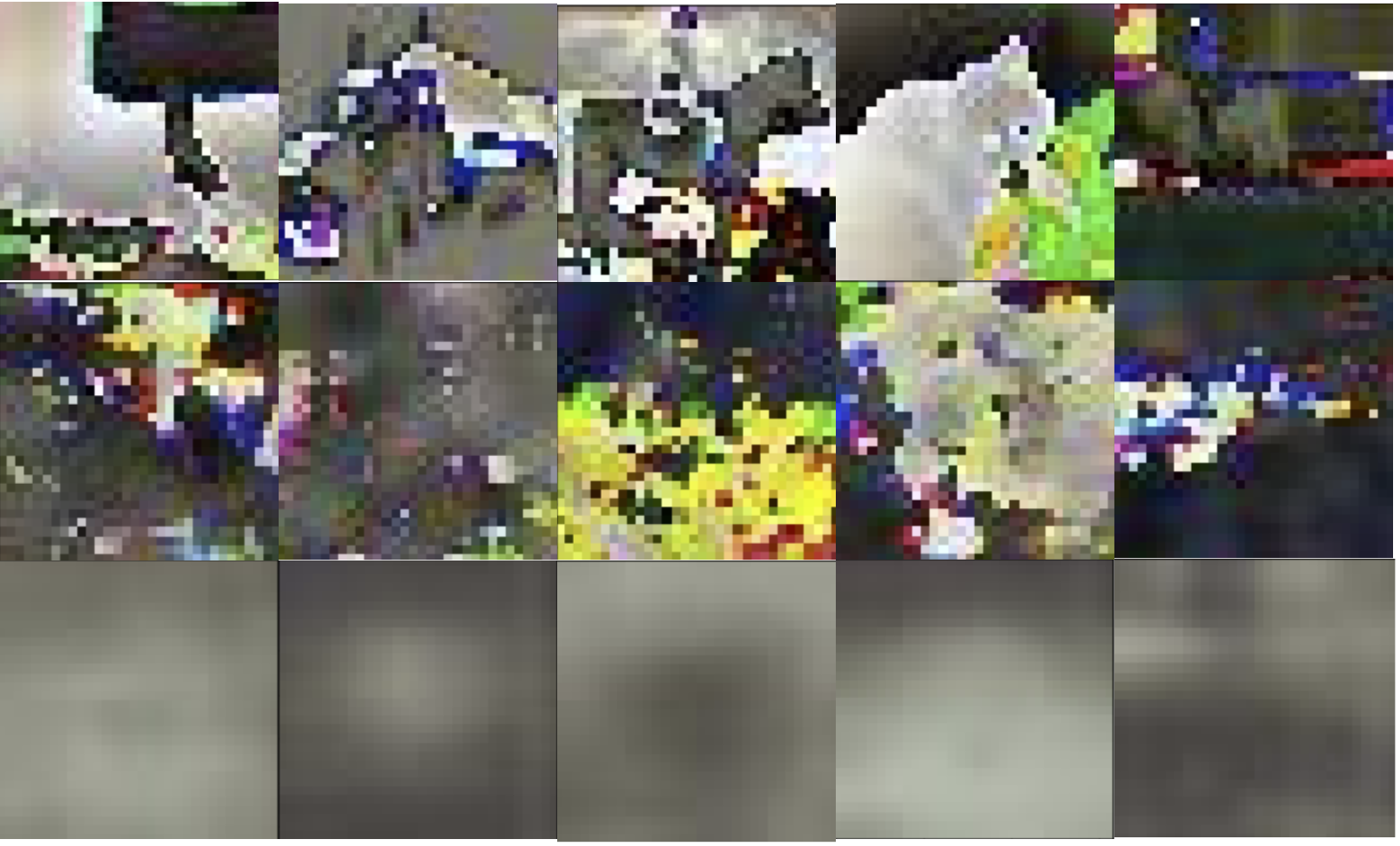}
  \caption{Reconstruction results for CIFAR10. The top row is the original image and the second row is reconstruction from activations of the network trained on CIFAR10. In the third row, activations are presented from network trained with NoPeek method. Even though the images are part of the training dataset, the network present in reconstruction is not able to generate any meaningful or discriminative representation when activations come from the network trained with NoPeek. The testbed fails to reconstruct the activations perfectly for the baseline as well especially because the activations belong to last layer.}
   \label{fig:cifar10train}
\end{figure}
 We choose one of the middle layers of ResNet-18, that is positioned at second stage and first residual block in the network. We train the ResNet-18 with both regular training (baseline) and NoPeek approaches. We then use this layer's activations on the validation as the input dataset for training the reconstruction testbed. The result is shown in the Figure~\ref{fig:cifar10train}. The low resolution of images in the CIFAR10 dataset makes it difficult to report the results experimentally. However, degradation in the reconstruction quality is clearly evident in the Figure~\ref{fig:cifar10train}.

\subsection{UTKFace}
UTKFace is a database of human faces. We train a ResNet based branched network which learns to predict the age, race, and gender of the person. The initial part of the network is common for all three prediction branches and splits near the end of the network with dense layers. We use ResNet-18 for the common part of the architecture and choose the last ResNet block of the output of second stage to be decorrelated with the input for testing the NoPeek approach.

\begin{figure}
    \centering
    \includegraphics[scale=0.27]{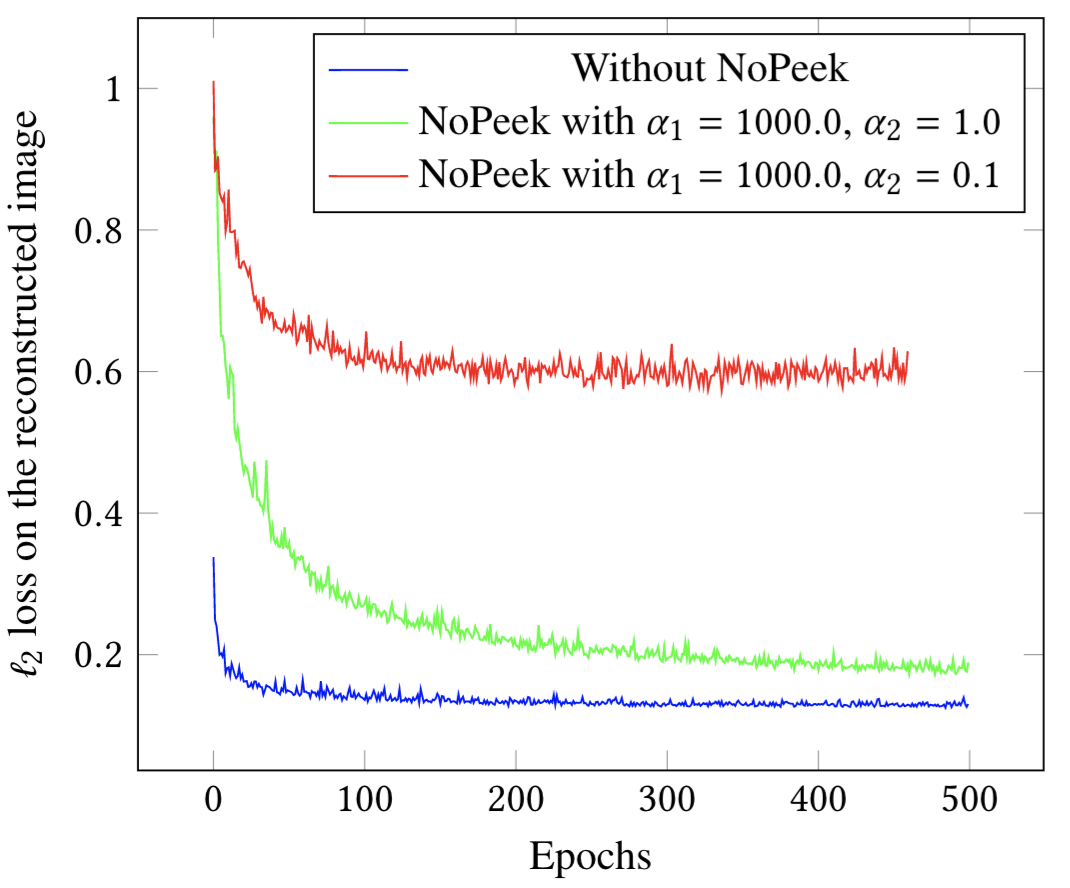}
        \caption{average $\ell_2$ norm between the image reconstructed by reconstruction testbed and original image in UTKFace dataset. Changing $\alpha$ results in different levels of the difference in the . $\ell_2$ loss between the two images is not the ideal metric as it does not handle the semantic features present in the image. For the qualitative results please see Figure 8.}
    \label{fig:l2_test}
\end{figure}

Figure~\ref{fig:faces} shows the qualitative result of our experiment on the face attribute prediction. We observe majority of the faces to be unidentifiable from the reconstruction.
 \begin{figure}
    \centering
   
    \includegraphics[scale = 0.10]{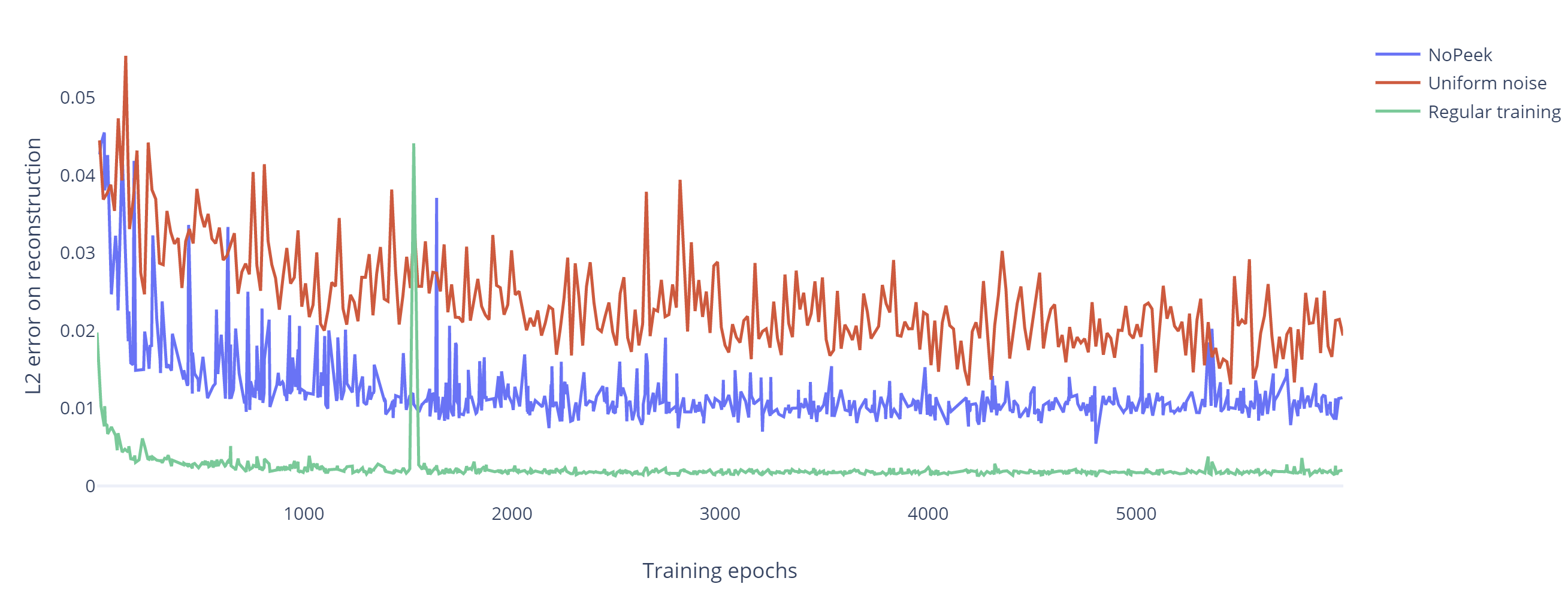}
    
    \caption{\textbf{Privacy-utility tradeoff on UTK:} We show $l_2$ error of reconstruction of a baseline strategy of adding uniform noise (in red) to activations of the layer being protected. This results in a model of no classififcation utility (performs at chance accuracy) albeit while preventing reconstruction. Our NoPeek approach (in blue) attains a much greater classfication accuracy for the downstream task (~0.82) compared to adding uniform noise (~chance accuracy) while still preventing reconstruction of raw data. This is compared to regular training, that does not prevent the reconstruction (in green).}
    \label{l2recons_utk_noise}
    \end{figure}
   
In general, we observe the reconstructions learned by the testbed trained on NoPeek activations tend to be relatively similar towards average face image of this dataset. For quantitative comparison, we plot the average $l_2$ reconstruction error for the entire test dataset in Figure~\ref{fig:l2_test}. In Figure \ref{l2recons_utk_noise}. we show the privacy-utility tradeoff with respect to a baseline of adding uniform noise, NoPeek and conventional training.

\begin{figure}
    \centering
    \includegraphics[scale=0.5]{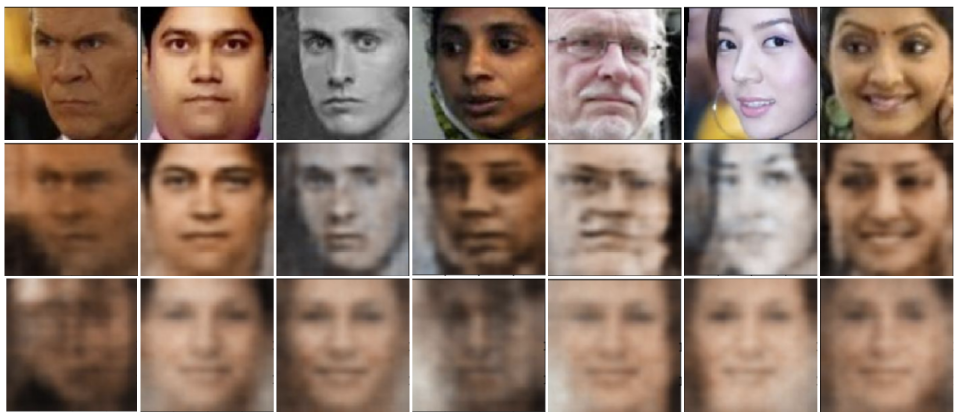}
    \caption{We use the reconstruction testbed to generate faces from the activations of a given intermediate layer. Here, first row is the actual image, second row is reconstruction from the activations and third row is reconstruction when the network is trained with NoPeek. NoPeek training makes it difficult for the adversary to generate the actual image from the activations.}
    \label{fig:faces}
\end{figure}

\subsection{Diabetic Retinopathy}
\begin{figure}
    \centering
    \includegraphics[scale=0.12]{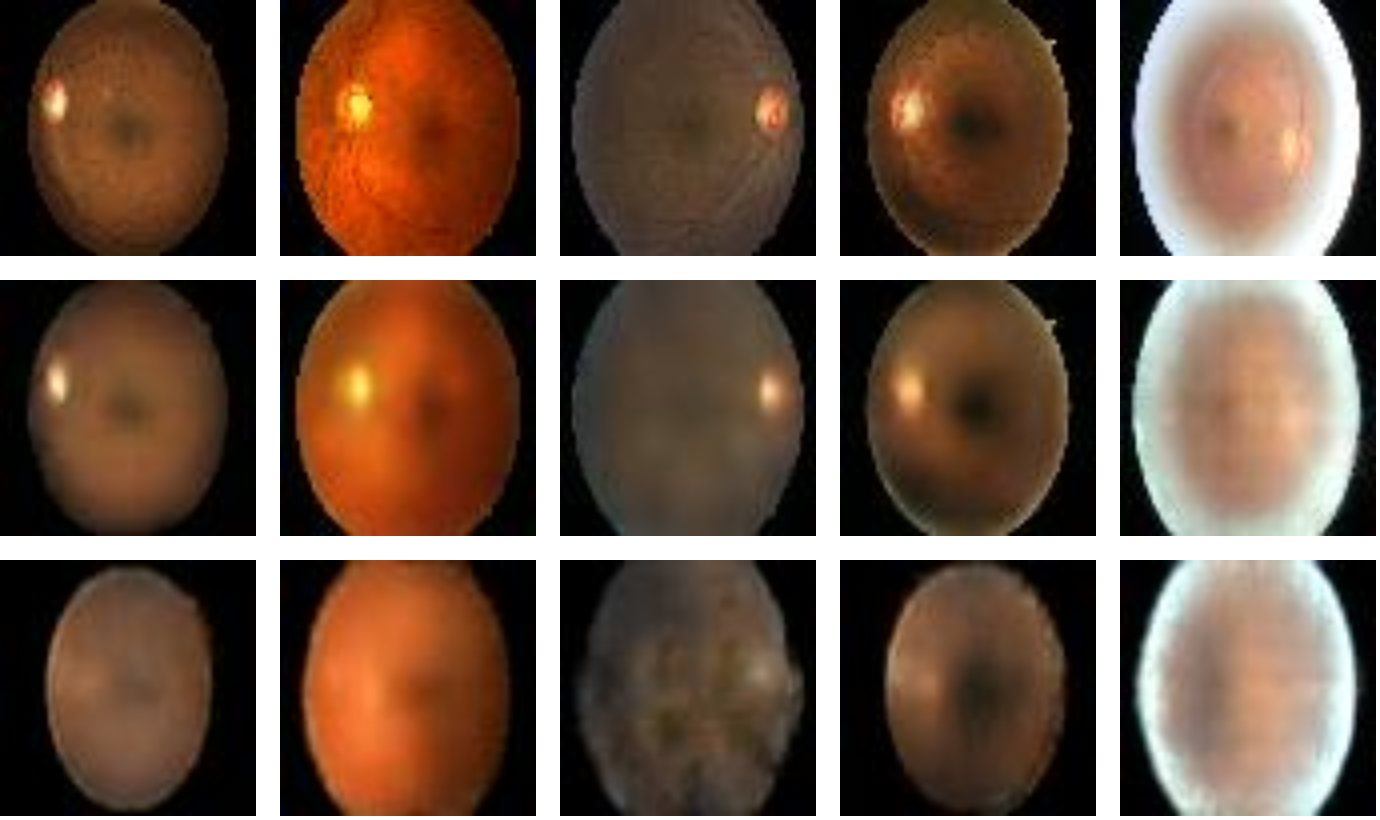}
    \caption{Reconstruction results for fundus images with first row being the original image, second row as reconstruction from regular training and third is the result of reconstruction from NoPeek. The finer level granularity of vessels and dark spots is lost in both images but the no-peek approach loses it even more making better for privacy of attributes which can be inferred from these finer level details which is needed for diagnosis or biometric applications. }
    \label{fig:dr}
\end{figure}

Privacy is a well known concern in the medical community hence, we experiment NoPeek approach for training network on the task of Diabetic retinopathy severity detection method. Previous research~\cite{Poplin2018} has shown it is possible to predict personal attributes of a person like gender, smoking habits etc. from fundus images. In our experiment, we train a CNN model to predict diabetic retinopathy severity from the fundus images. We use standard ResNet-18 for training the main model and partition the activations in the middle of the layers. We downsample the fundus images to standardize them to a common size of $64 \times 64$. Hence, the task for the reconstruction testbed is to generate $64 \times 64$ fundus image given the intermediate activations. Figure~\ref{fig:dr} shows qualitative results for some of the samples, it can be noted that the images generated by the testbed do not reconstruct attribute discriminative features such as blood vessels successfuly and the loss of such discriminative features is higher for the NoPeek method.

\begin{figure}
    \centering
    \includegraphics[scale=0.27]{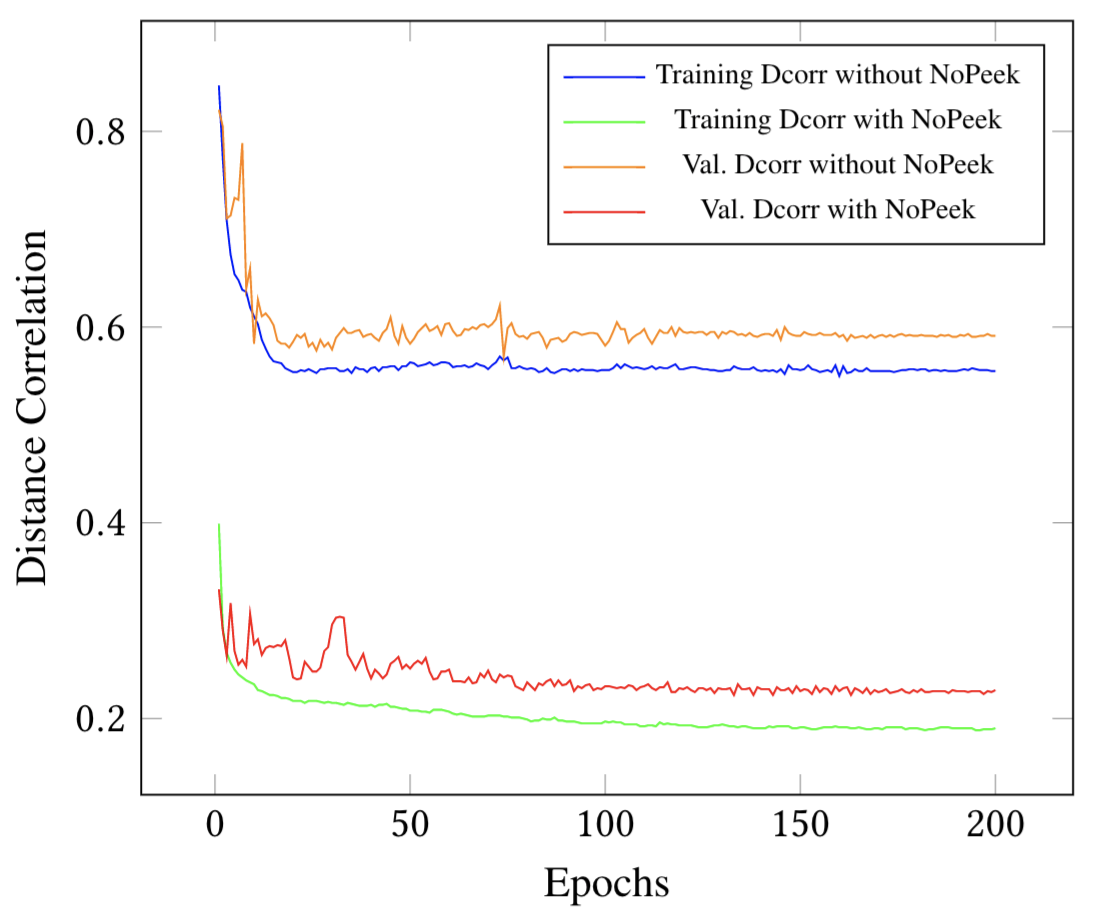}
    \caption{Distance correlation value as the network gets trained on the UTK face dataset. Note that the network by itself reduces distance correlation even in the baseline experiment.}
    \label{plot:1}
\end{figure}
\subsection{Attribute Privacy}
As described mathematically in the section 4.2, we also extend the NoPeek approach to minimize distance correlation between the intermediate activations and a particular chosen attribute with respect to which we want to attain privacy. We use UTK Face dataset under the same setting as described in the section 7.2, but this time instead of training the method on all three attributes - age, gender, and race, we only train on the two attributes at a time while treating the other attribute as a protected class and hence, distance correlation is minimized between the intermediate activations and the corresponding attribute. 
\begin{figure}[!htbp]
  \centering
  \includegraphics[scale=0.11]{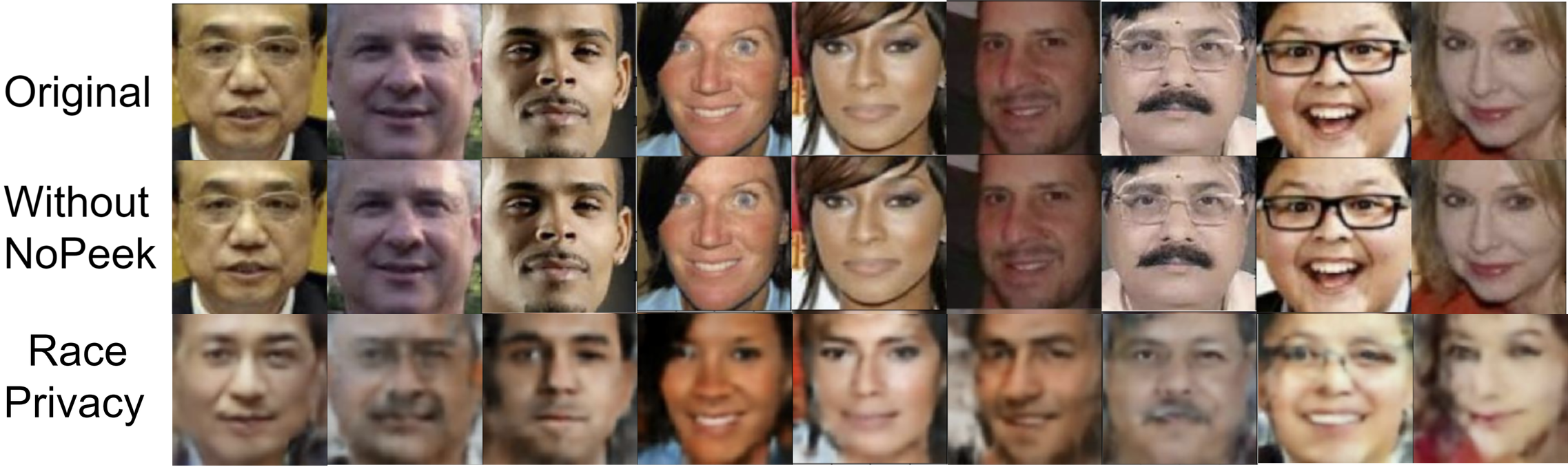}
  \caption{We extend the NoPeek approach to decorrelate the activations with respect to a particular attribute. In this figure, we decorrelate the intermediate activations with the attribute (race) we want to protect from leaking during inference.  Usually, the information leakage is still very high in the activations upon applying only a single layer of convolution filters, yet, we can see here that the NoPeek approach makes it difficult for the adversary to reconstruct the person’s image.
  }
  \label{fig:age-attribute}
\end{figure}
We do not treat gender as a protected attribute since it is a binary class in our dataset. Figure~\ref{fig:age-attribute} shows the reconstruction results when NoPeek approach is used for attribute level privacy. The more compelling part of this experiment is that the reconstruction testbed is never supervised about the image's attributes itself, it is just designed to generate the image showing that attribute level privacy technique is indeed capturing related attributes and invariances.

\begin{figure}[htbp]
    \centering
    \includegraphics[scale=0.27]{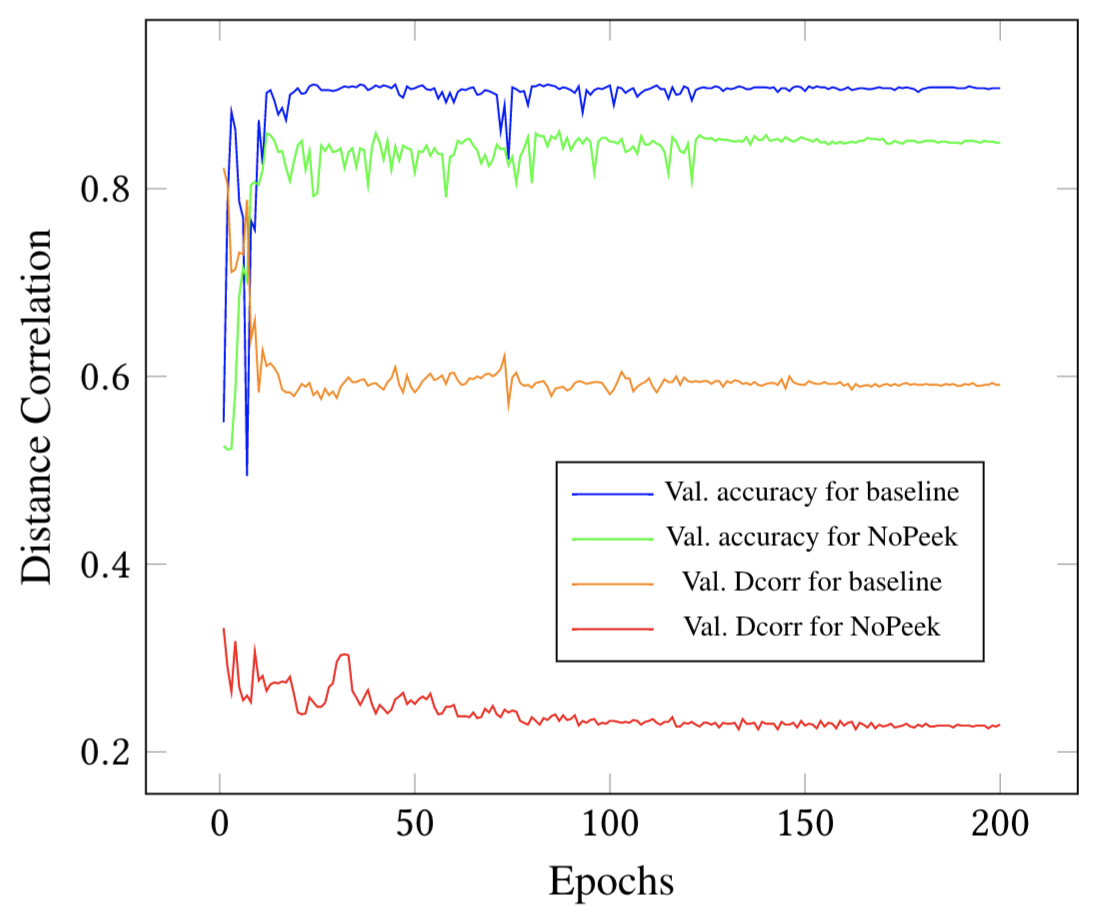}
    \caption{By introducing NoPeek in the training of the network, we obtain a major decrease in the distance correlation from 0.6 (baseline) to 0.22 (NoPeek) while the decrease in the accuracies is relatively much lesser.}
    \label{plot:2}
\end{figure}

\subsection{Visualizing activations}

\begin{figure}[htbp]
    \centering
    \includegraphics[scale=0.27]{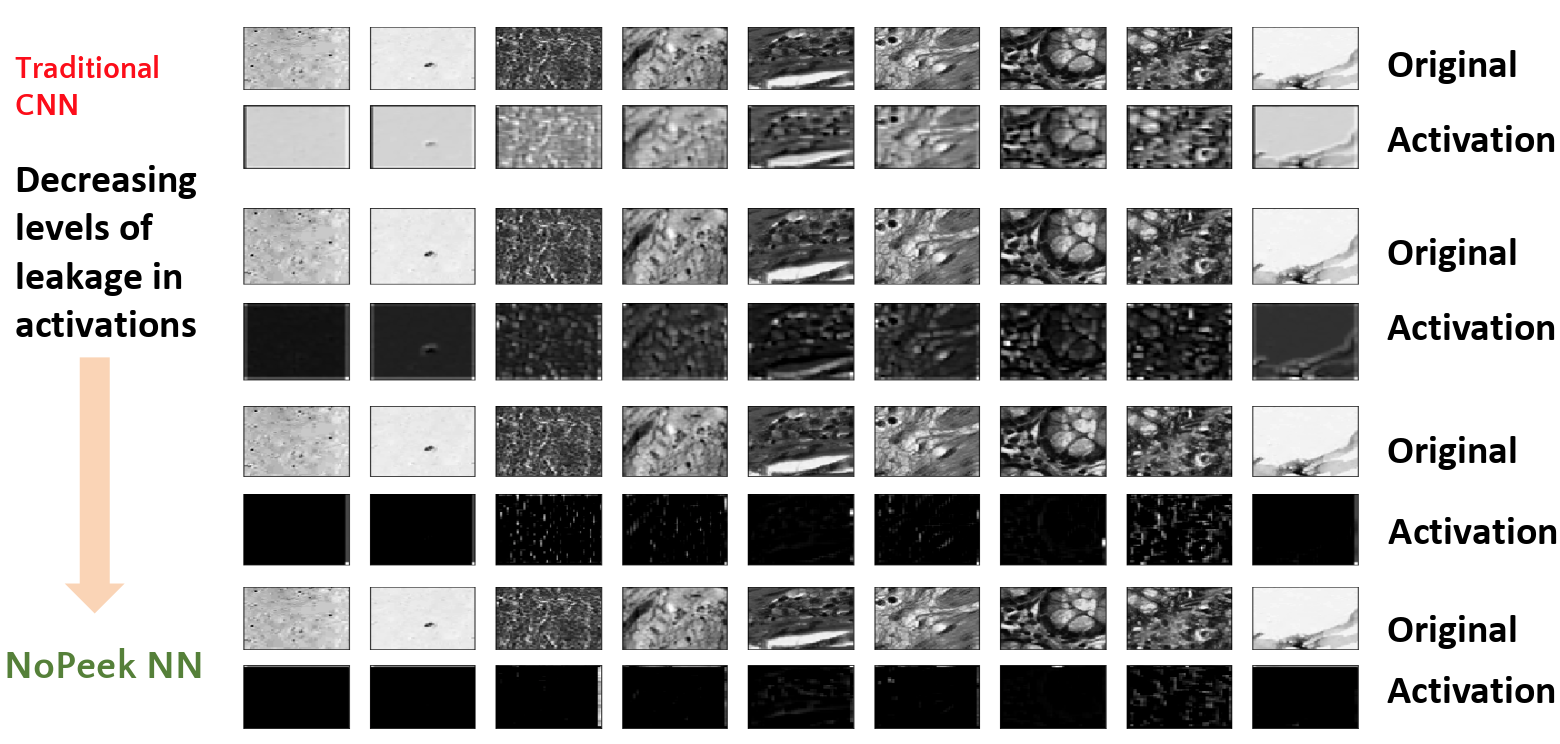}
    \caption{We see decreasing levels of leakage of information about raw data in the activations as the weight of distance correlation term in the weighted loss function is increased significantly over a colorectal histology medical dataset available publicly.}\label{FigColoac}
\end{figure}

We visualize the activations of the filters in early layers to see the resulting effect of minimizing the distance correlation between activations and raw data. Figure \ref{FigColoac} shows the decreasing levels of leakage.

\begin{wrapfigure}[23]{r}{0.21\textwidth}
  \centering
  \includegraphics[scale=0.07]{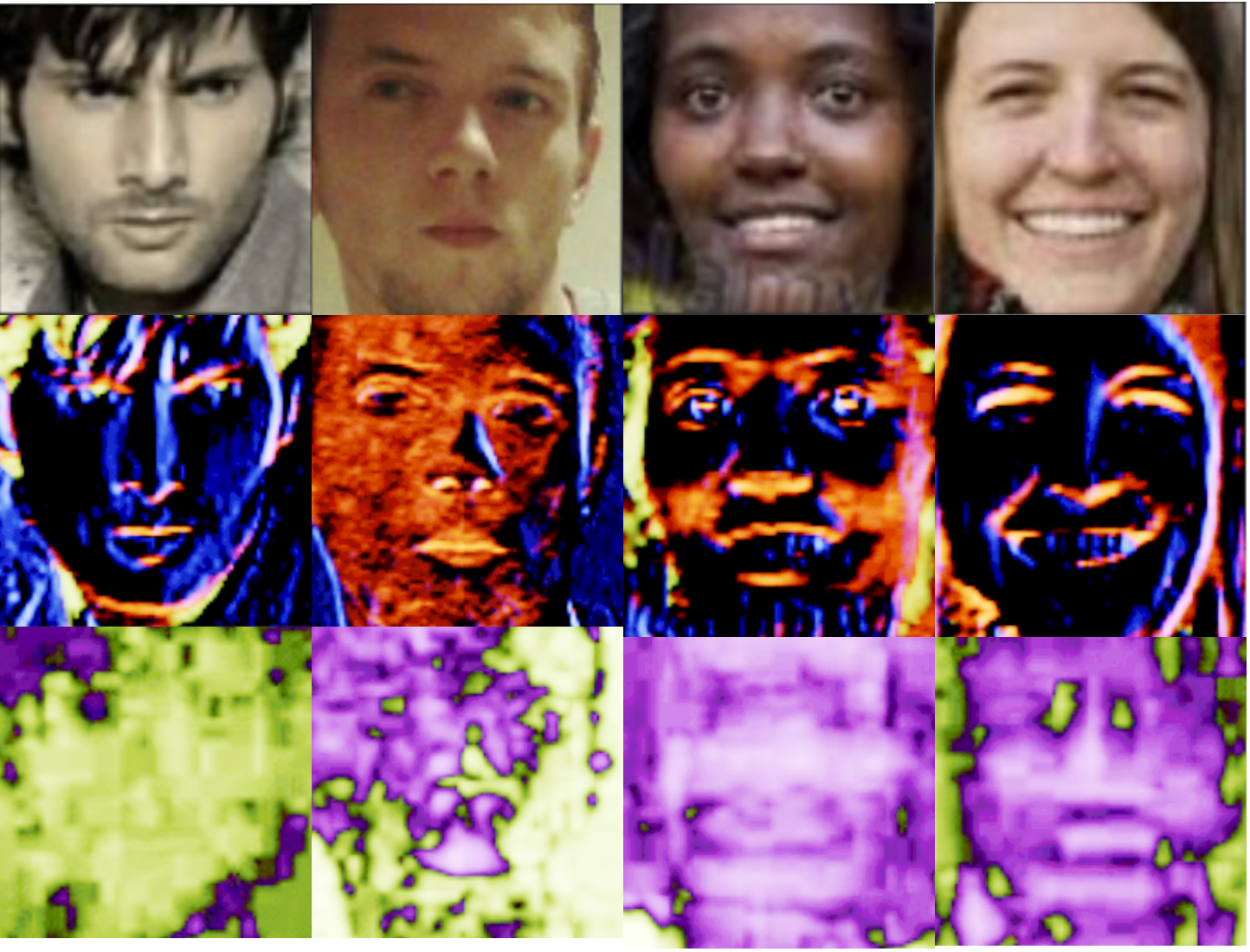}
  \caption{Visualization of the activations of the first layer of a ResNet. In the activation maps of the first layer in the second row, subtle facial features can be observed from the activations about the raw image while, in the third row, the NoPeek method forces the network to decorrelate the features with respect to raw data, hence making it hard to interpret.}
  \label{fig:activations}
\end{wrapfigure}

For this experiment, we treat the target $z$ for NoPeek as the output of first CNN layer itself which we restrict to only three output channels so as to visualize only the RGB component. Figure~\ref{fig:activations} shows the output of first layer of the trained network. The joint minimization of distance correlation with cross entropy(in classification task) leads to a different set of feature extraction or transformation over features in such a way that it is perceivable for human visual system as well.
We also visualize activations obtained on colorectal histology dataset with increasing values of $\alpha$ in \ref{FigColoac}.

\section{Discussion}
One of the important aspects of proposed technique is to jointly optimize for distance correlation and task related loss function like cross entropy for classification. In other words, we are optimizing for the trade-off between privacy and utility by controlling $\alpha_1$ and $\alpha_2$ as described previously. Figure~\ref{fig:l2_test} shows three variations of this trade-off , where \textit{Without NoPeek} approach is essentially $\alpha_1=0$. In order to understand it well, in Figure~\ref{plot:1} we plot the distance correlation of a fixed intermediate activation as during training for a NoPeek network as well as for a network without NoPeek. This demonstrates that the network without NoPeek also reduces the distance correlation beyond a certain layer and our proposed method can be seen as an additional regularizer which forces the network to regularize for the reduction in distance correlation at a much higher rate between raw data and the activations.
In Figure~\ref{plot:2} we observe that accuracy  dropped by a relatively small amount compared to the drop in distance correlation and this relative difference between the drop can be controlled through tuning the $\alpha$ for the distance correlation as well as cross entropy. The choice of $\alpha$ is also dictated by the position of the intermediate activation in the network as well as the type of layer which produces the output. For example, compared to convolution layers, which imposes heavy prior on images, a fully connected layer can learn to attain a lesser distance correlation relatively easy and hence should also guide the choice of $\alpha$.
\section{Conclusion}
The proposed NoPeek schemes based on distance correlation seem to have versatile applicability in the space of privacy, computer vision and machine learning given that it does not require major changes in the model setup and architectures except for the proposed modification to loss function.It would be great to realize on-device implementations of the universal decorrelation and other NoPeek schemes. With regards to human visual perception of bias and privacy, we would also like to conduct a large-scale crowdsourced survey to compare performance of human participants in deciphering the true sensitive attribute upon looking at NoPeek results in terms of their proximity to a uniform random prediction.




%

\appendix
\section{A: Gradient of distance correlation}\label{gradDC}
Distance correlation between centered data can be represented as $\frac{\Tr{(\mathbf{X^TXZ^TZ})}}{\sqrt{\Tr{(\mathbf{X^TX})^2}\Tr{(\mathbf{Z^TZ})^2}}}$ \cite{vepakomma2018supervised}. Distance covariance in the numerator can be written as 
\begin{equation}
    \Tr(\mathbf{X^{T}ZX}) =\sum_{ij}\langle z_i,z_j \rangle (\lVert x_i-x_j\rVert)^2
\end{equation}
This can be written in matrix form using basis vectors $e_i,e_j$ as 
\begin{equation}\label{matNot}
    \sum_{ij}[\Tr\mathbf{(Z^Te_ie_j^TZ)Tr(X^T(e_i-e_j)(e_i-e_j)^TX)}]
\end{equation}
Simplifying the notation with $M_{ij} = e_ie_j^T$ and $A_{ij} = (e_i-e_j)(e_i-e_j)^T$ we have 
$$\frac{\partial Tr (\mathbf{Z^TL_ZZ}) }{\partial \mathbf{Z}} = \sum_{ij}\mathbf{(2M_{ij}Z)Tr(X^TA_{ij}X)}$$
On the lines of \ref{matNot}, we have $Tr(\mathbf{Z^TL_ZZ}) = \sum_{ij}[Tr\mathbf{(Z^TM_{ij}Z)Tr(Z^TA_{ij}Z)}]$
Therefore utilizing these identities, the derivative of squared distance correlation w.r.t $\mathbf{Z}$ can be written as 
\begin{equation}
   \frac{ c_x Tr(\mathbf{Z^TL_Z Z})\frac{\partial Tr (\mathbf{X^TL_ZX}) }{\partial \mathbf{Z}}-[Tr(\mathbf{X^TL_ZX})]^2 c_x \frac{\partial Tr (\mathbf{\mathbf{Z^TL_ZZ}}) }{\partial \mathbf{Z}}}{[Tr(\mathbf{Z^TL_ZZ})]^2}
\end{equation}
\subsection{B: Deep-learning friendly source code for sample distance correlation}
\begin{lstlisting}[language=Python]
def pairwise_dist(A):
    r = tf.reduce_sum(A*A, 1)
    r = tf.reshape(r, [-1, 1])
    D = tf.maximum(r - 2*tf.matmul(A, tf.transpose(A)) + tf.transpose(r), 1e-7)
    D = tf.sqrt(D)
    return D

def dist_corr(X, Y):
    n = tf.cast(tf.shape(X)[0], tf.float32)
    a = pairwise_dist(X)
    b = pairwise_dist(Y)
    A = a - tf.reduce_mean(a, axis=1) -\
    tf.expand_dims(tf.reduce_mean(a,axis=0),axis=1)+\
    tf.reduce_mean(a)
    B = b - tf.reduce_mean(b, axis=1) -\
    tf.expand_dims(tf.reduce_mean(b,axis=0),axis=1)+\
    tf.reduce_mean(b)
    dCovXY = tf.sqrt(tf.reduce_sum(A*B) / (n ** 2))
    dVarXX = tf.sqrt(tf.reduce_sum(A*A) / (n ** 2))
    dVarYY = tf.sqrt(tf.reduce_sum(B*B) / (n ** 2))
    
    dCorXY = dCovXY / tf.sqrt(dVarXX * dVarYY)
    return dCorXY
\end{lstlisting}

\subsection{\textbf{Code/Reproducibility}} The code for our method is provided in this anonymous code repository: \url{https://anonymous.4open.science/r/820473f8-f3ee-4212-9b9c-409a78722af6/}.

\bibliographystyle{IEEEtranN}
\bibliography{bare_conf}

\end{document}

%% file: math_commands.tex
\usepackage{amsmath,amsfonts,bm}

\def\eqref#1{equation~\ref{#1}}

\def\1{\bm{1}}

\DeclareMathAlphabet{\mathsfit}{\encodingdefault}{\sfdefault}{m}{sl}
\SetMathAlphabet{\mathsfit}{bold}{\encodingdefault}{\sfdefault}{bx}{n}

